\documentclass{article}

     \PassOptionsToPackage{numbers, compress}{natbib}

     \usepackage[preprint]{neurips_2020}

\usepackage[utf8]{inputenc} 
\usepackage[T1]{fontenc}    
\usepackage{hyperref}       
\usepackage{url}            
\usepackage{booktabs}       
\usepackage{amsfonts}       
\usepackage{nicefrac}       
\usepackage{microtype}      

\usepackage{soul}
\usepackage{url}

\usepackage[small]{caption}
\usepackage{graphicx}
\usepackage{subfigure}
\usepackage{amsmath}
\usepackage{epstopdf}
\usepackage{amssymb}
\usepackage{algorithm}
\usepackage{algorithmic}
\def\VectorFont{\bf}

\newcommand{\vw}{{\VectorFont w}}
\title{Optimally Combining Classifiers for Semi-Supervised Learning}

\author{
Zhiguo Wang$^{1,2,3}$~~
Liusha Yang$^3$~~
Feng Yin$^{1,3}$~~
Ke Lin$^{4}$~~
Qingjiang Shi$^{5,3}$~~
Zhi-Quan Luo$^{1,3}$
\thanks{The work of Zhi-Quan Luo is supported by the leading talents of Guangdong province Program (No. 00201501), the National Natural Science Foundation of China (No. 61731018), Shenzhen Peacock Plan (No.  KQTD2015033114415450), the Development and Reform Commission of Shenzhen Municipality, and Shenzhen Research Institute of Big Data.}
\\
$^1$School of Science \& Engineering, The Chinese University of Hong Kong, Shenzhen 518172, China\\
$^2$University of Science and Technology of China, China\\
$^3$Shenzhen Research Institute of Big Data, Shenzhen 518172, China\\
$^4$Huawei Technologies Co., Ltd\\
$^5$School of Software Engineering, Tongji University, Shanghai 200092, China\\
\texttt{
wangzhiguo@cuhk.edu.cn,yangliusha@sribd.cn, yinfeng@cuhk.edu.cn,
}\\
\texttt{linke2@huawei.com,
shiqj@tongji.edu.cn,luozq@cuhk.edu.cn,}
}

\begin{document}

\maketitle

\begin{abstract}
  This paper considers semi-supervised learning for tabular data. It is widely known that  Xgboost based on tree model works well on the heterogeneous features while transductive support vector machine can exploit the low density separation assumption. However, little work has been done to combine them together for the end-to-end semi-supervised learning. In this paper, we find these two methods have complementary
properties and larger diversity, which motivates us to propose a new semi-supervised learning method that is able to adaptively combine the strengths of Xgboost and transductive support vector machine. Instead of the majority vote rule, an optimization problem in terms of ensemble weight is established, which helps to obtain more accurate pseudo labels for unlabeled data. The experimental results on the UCI data sets and real commercial
data set demonstrate the superior classification performance of our method over the five state-of-the-art algorithms improving test accuracy by about $3\%-4\%$. The code can be found at https://github.com/hav-cam-mit/CTO.
\end{abstract}

\section{Introduction}
In many applications, it is a difficult task to obtain fully labeled data sets to train a classifier, and labeling is usually expensive, time consuming and subject to human expertise, yet collecting abundant unlabeled data is much easier \cite{cohen2004}. To leverage both labeled and unlabeled samples, semi-supervised learning (SSL) has been proposed to improve the generalization ability  \cite{zhu2005,chapelle2009}.

There are four broad categories of semi-supervised learning methods, i.e. generative
methods, graph-based methods, low-density separation
methods and disagreement-based methods \cite{zhou2005tri} to be discussed in Section \ref{sec2}. The above methods are proposed based on certain assumptions on the labeled data and unlabeled data, which plays an important role in semi-supervised learning.  However, it remains an open question on how to make the right assumptions on a real data.

\begin{figure}[t]
\centering
\subfigure[cjs]{
\hspace{-0.6cm}
\begin{minipage}[t]{0.5\linewidth}
\centering
\includegraphics[height=2.1in]{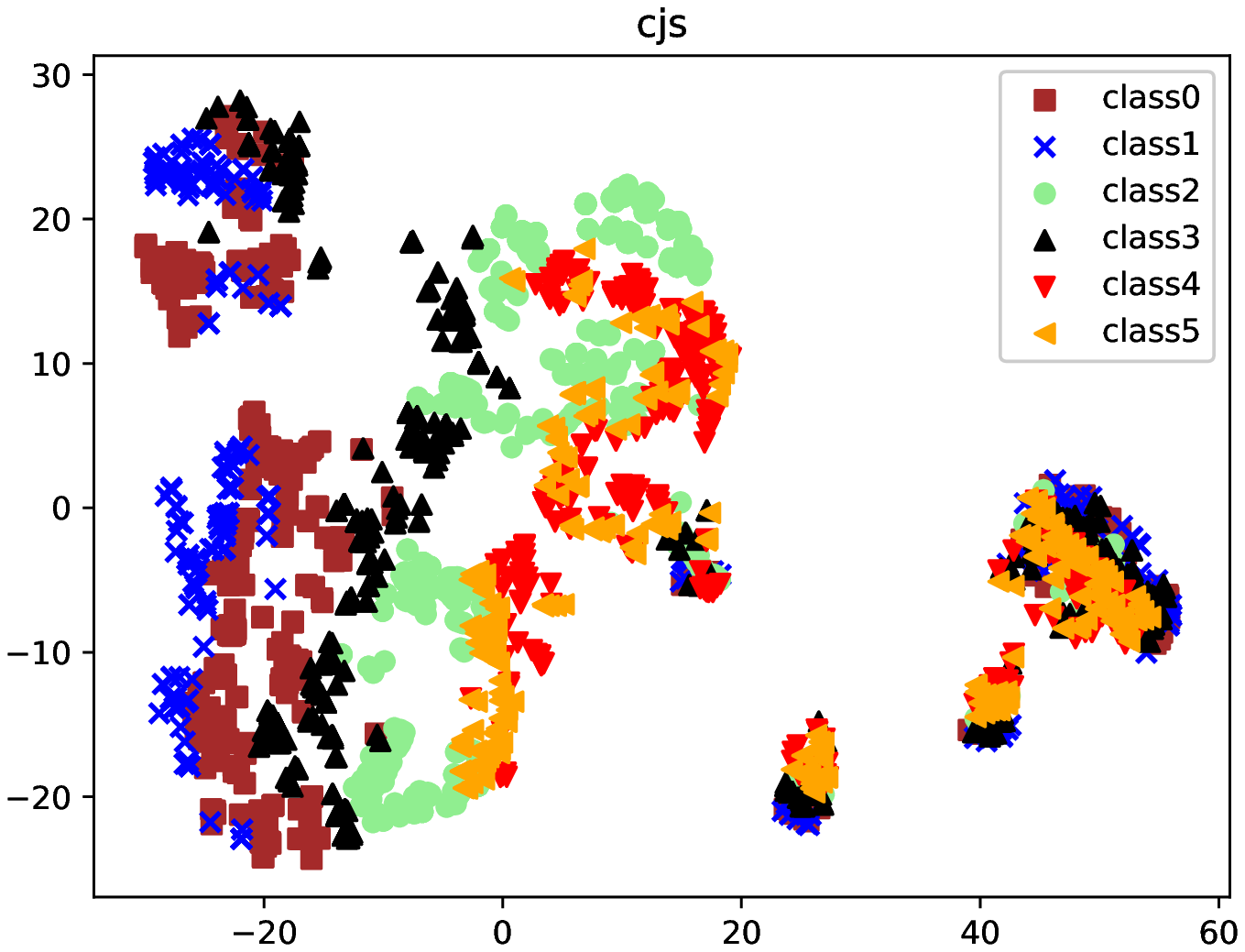}
\end{minipage}%
}%
\subfigure[analcat]{
\hspace{-0.15cm}
\begin{minipage}[t]{0.5\linewidth}
\centering
\includegraphics[height=2.1in]{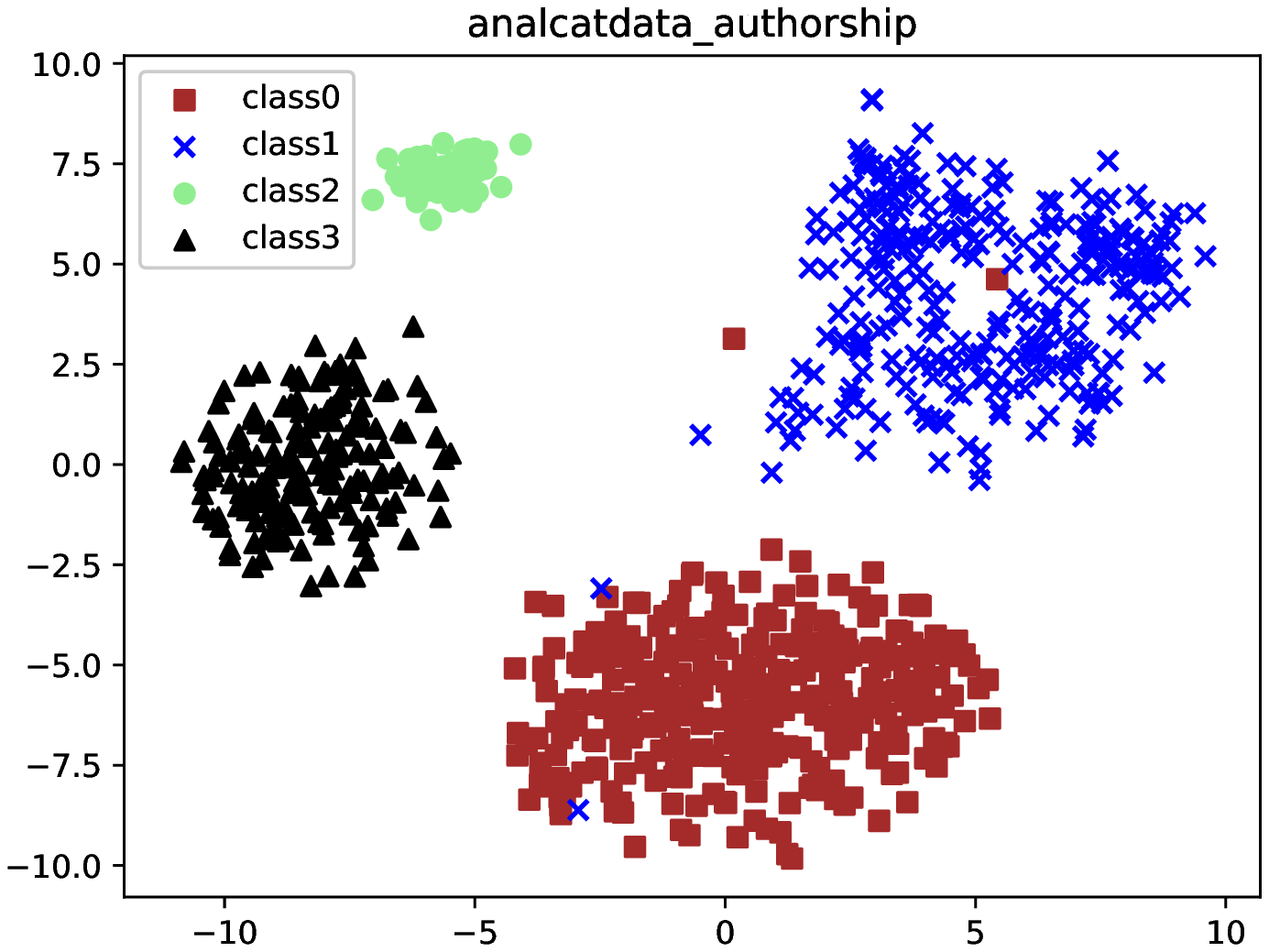}
\end{minipage}%
}%
\vspace{-0.2cm}
\centering
\caption{ Visualization of two data sets. (a) The test accuracy of TSVM and co-forest on cjs data is $0.654$ and $0.989$, respectively; (b) The test accuracy of TSVM and co-forest on analcat is $0.992$ and $0.876$, respectively.}
\label{motivation}
\end{figure}

By t-distributed stochastic neighbor embedding (t-SNE) tool  \cite{maaten2008}, Fig. \ref{motivation} visualizes two real data sets, namely the \emph{cjs} and \emph{analcat} data sets, which are downloaded from UCI repository. The test accuracy of the transductive support vector machine (TSVM) \cite{bennett1999} on the \emph{analcat} data set is better than that of co-forest \cite{li2007improve} model. But co-forest performs better  on \emph{cjs} data. The reason may be that \emph{analcat} has large margin between classes such that the assumption of low-density separation is satisfied for TSVM (see Fig. \ref{motivation}). For \emph{cjs} data, the class distribution is irregular and tree based method can work well in this situation. Seemingly, TSVM and tree based model have complementary properties.  This motivates us to ensemble heterogeneous classifiers for semi-supervised learning.
In this paper, we propose a new semi-supervised method, called co-training with optimal weight (e.g., CTOW).  The contributions of our method are in order.
\begin{itemize}
  \item We combine co-training with two strong heterogeneous classifiers, namely, Xgboost and TSVM, which have
  complementary properties and larger diversity.
  \item The optimization problem of the weight for each classifier is established and we provide prior information of the margin density to help compute the weight of TSVM.
  \item The proposed method works well for both large margin data (e.g., \emph{analcat} data set) and irregular data (e.g., \emph{cjs} data set).
\end{itemize}
Experiments are conducted on fourteen real tabular data sets. The results show that our method can improve at least $3\%$ test accuracy with less computational time.
\section{Related Work}\label{sec2}


Generative methods \cite{miller1997mixture} are simple and effective in semi-supervised learning. The assumption  about generative methods is that data actually comes from a mixture model \cite{zhu2005}. Graph-based methods \cite{zhu2003semi} use similarity matrix to construct a graph. The main assumption made by graph-based methods is that the labels are smooth with respect to the graph. However, the efficiency of the graph-based methods will heavily depend on the size of the constructed graph.

Low-density separation methods assume that the classes are well-separated, such that the decision boundary lies in a low-density region and does not cut through dense unlabeled data region. The most famous representative among others is semi-supervised support vector machine (S3VM), also called TSVM. The optimization problem of TSVM is given as follows \cite{bennett1999}
\begin{align}
\label{TSVM1}{\min _{\boldsymbol{w}, b, \hat{y}, \xi}} & {\frac{1}{2}\|\boldsymbol{w}\|_{2}^{2}+C_{l} \sum_{i=1}^{l} \xi_{i}+C_{u} \sum_{j=l+1}^{l+u} \xi_{j}}\\
{\text { s.t. }} & {y_{i}\left(\boldsymbol{w}^{\mathrm{T}} x_{i}+b\right) \geqslant 1-\xi_{i},~\xi_{i} \geqslant 0 \quad i=1,2,\ldots,l} \\
\label{TSVM2}& {\hat{y}_{j}\left(\boldsymbol{w}^{\mathrm{T}} x_{j}+b\right) \geqslant 1-\xi_{j},
 \hat{y}_{j}\in\{-1,1\},~\xi_{j} \geqslant 0, \quad j=l+1,\ldots,l+u} ,
\end{align}
where $\boldsymbol{w},b$ are the parameters that specify the orientation and the offset, respectively; $\xi$ is the slack variable; $\hat{y}$ is the pseudo label to be optimized; $C_l$ and $C_u$ are the penalty constants; the set $\{x_i,y_i\}_{i=1}^l$ represents the labeled data, and the set $\{x_j\}_{l+1}^{l+u}$ represents the unlabeled data.
Since \eqref{TSVM1}-\eqref{TSVM2} is a non-convex optimization problem, many researchers strived to solve it efficiently \cite{chapelle2008}.

Disagreement-based methods \cite{blum1998,zhou2010} need assemble of multiple learners and let them collaborate and teach each other to
exploit unlabeled data. Co-forest \cite{li2007improve} is one of the most famous representative methods, which extends the co-training paradigm \cite{blum1998} by random forest consisting of many trees. Each decision tree is firstly initiated from the training sets, then the unlabeled examples are randomly selected to label in confidently, finally, majority voting is employed to obtain the pseudo labels.
But co-forest is based on the ensemble of the weak classifiers. It is more desirable to exploit heterogeneous ensemble of strong classifiers with complementary properties to improve the performance of a co-training method.

Many recent approaches for semi-supervised learning advocate to train a neural network based on the consistency loss, which forces the model to generate consistent outputs when its inputs are perturbed, such as pseudo-labeling \cite{lee2013pseudo}, Ladder network \cite{rasmus2015}, $\Pi$ model \cite{laine2016}, mean teacher \cite{tarvainen2017}, VAT \cite{miyato2018virtual}, Mixmatch \cite{berthelot2019mixmatch}. The consistency assumption  works well for image data, among others video oriented tasks.  Nevertheless, the neural network with consistency assumption may not reflect any specific inductive bias toward tabular data.

So far gradient boosting decision trees \cite{friedman2001} and Xgboost \cite{chen2016xgboost} are two most widely used models in Kaggle competitions. Xgboost is an additive tree ensemble model aggregating outputs of $D$ trees
according to
$
\tilde{y}_i=\sum_{d=1}^Df_d(x_i),
$
where each $f_d$ is a regression tree, and $\tilde{y}_i$ is the final output for the input data $x_i$. To learn  the model parameter, they minimize the following loss function,
$
\mathcal{L} = \sum_{i=1}^l l(\tilde{y}_i,y_i)+\sum_d \Omega(f_d),
$
where $\sum_d \Omega(f_d)$ is a regularized term. 
As we known, Xgboost works well for tabular data, but  it only use the labeled data without unlabeled data.

Our paper centers around the semi-supervised learning for tabular data. The above discussions about Xgboost  inspires us to incorporate Xgboost classifier into semi-supervised learning.  In \cite{mallapragada2008}, the authors proposed method called semi-boost, which combines similarity matrix with boosting methods to obtain more accurate pseudo labels. However, semi-boost is computationally expensive for large data set. Thus, our paper uses disagreement-based method by the ensemble of Xgboost and TSVM to surpass  state-of-the-art  performance in semi-supervised learning.
\section{The Proposed Approach}\label{sec3}
\begin{figure}[t]
\begin{center}
\includegraphics[width=0.8\linewidth]{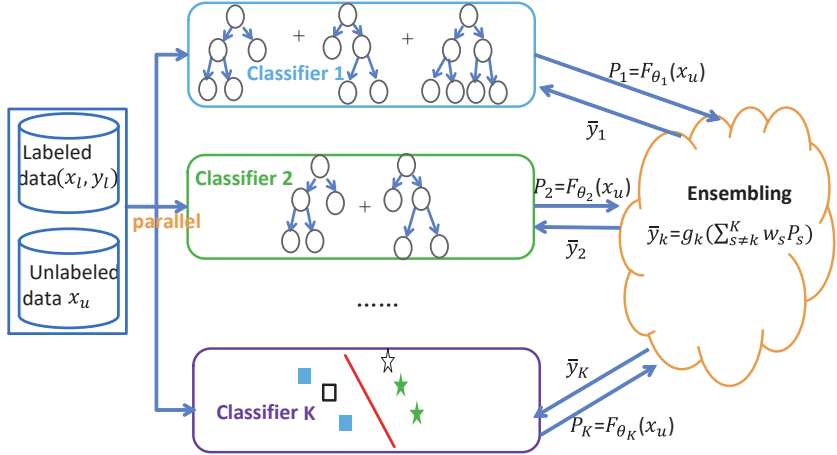}
\caption{The architecture of the proposed method, which combines Xgboost and TSVM.}
\label{fig:architecture}
\end{center}
\end{figure}

In this paper, we consider semi-supervised classification problems. The training set consists of a labeled data set $\mathcal{L}=\{(x_i,y_i)\}_{i=1}^l$ with $l$ labeled examples and $u$ unlabeled examples $\mathcal{U}=\{x_i\}_{i=l+1}^{l+u}$, with $l\ll u$. Assume that the data has $C$ classes. We attempt to utilize training set $\mathcal{L}\cup\mathcal{U}$ to construct a learner to classify unseen instances. To this end, we propose a new semi-supervised learning method CTOW, by
combining the Xgboost with  TSVM. The reason that selecting Xgboost and  TSVM as base learner is explained in Subsection \ref{MD}.  The architecture of our method is shown in Fig. \ref{fig:architecture}, and the detailed techniques are presented in the following subsection.
\subsection{Select Base Learners}\label{MD}
There are two directions to select the base learners for semi-supervised learning, one is diversity of the learners,  which is deemed to be a key of good ensemble \cite{Zhou2010Semi}. The other is the accuracy of the learners, which helps us to find some better pseudo label data and improve the classification accuracy further.

To select some base learners maintaining a large diversity, we should use a criteria to measure the diversity. As we know, the correlation coefficient $\rho$ is a simple and efficient method to measure the diversity, which is the correlation between two classifier outputs (correct/incorrect) \cite{Kuncheva2003Measures}. It is formulated as follows
\begin{align}
\label{eqn: rho} \rho=\frac{N^{11} N^{00}-N^{01} N^{10}}{\sqrt{\left(N^{11}+N^{10}\right)\left(N^{01}+N^{00}\right)\left(N^{11}+N^{01}\right)
\left(N^{10}+N^{00}\right)}},
\end{align}
where $N^{11}$, $N^{10}$, $N^{01}$ and $N^{00}$ are defined in Table \ref{tab: correlation}.
\begin{table}[t]
\centering
\caption{The relationship between a pair of classifiers}
\label{tab: correlation}
\begin{tabular}{lcc}
\hline & Classifier one correct & Classifier one wrong \\
\hline Classifier two correct & $N^{11}$ & $N^{10}$ \\
Classifier two wrong  & $N^{01}$ & $N^{00}$ \\\hline
\end{tabular}
\end{table}
According to \eqref{eqn: rho}, we calculate the correlation coefficient $\rho_{TT}$ between two different decision tree classifiers for different real data. In addition, the correlation coefficient $\rho_{ST}$ between TSVM and decision tree classifier is given. We also show the the correlation coefficient $\rho_{SX}$ between TSVM and Xgboost.

\begin{figure}[t]
\centering
\subfigure[Measure of diversity]{
\hspace{-0.6cm}
\begin{minipage}[t]{0.5\linewidth}
\centering
\includegraphics[height=2.1in]{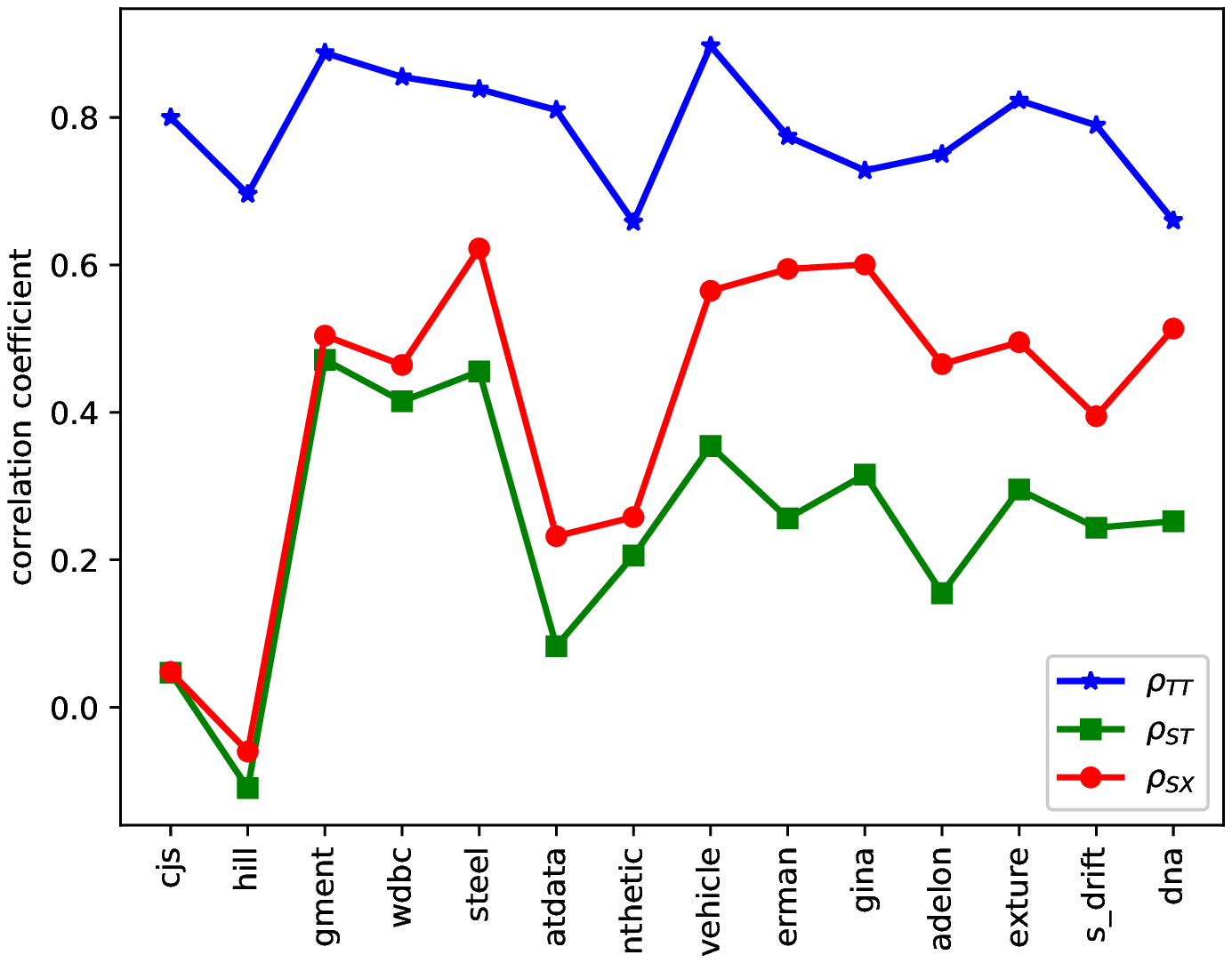}
\end{minipage}%
}%
\subfigure[Accuracy of classifiers]{
\hspace{-0.15cm}
\begin{minipage}[t]{0.5\linewidth}
\centering
\includegraphics[height=2.1in]{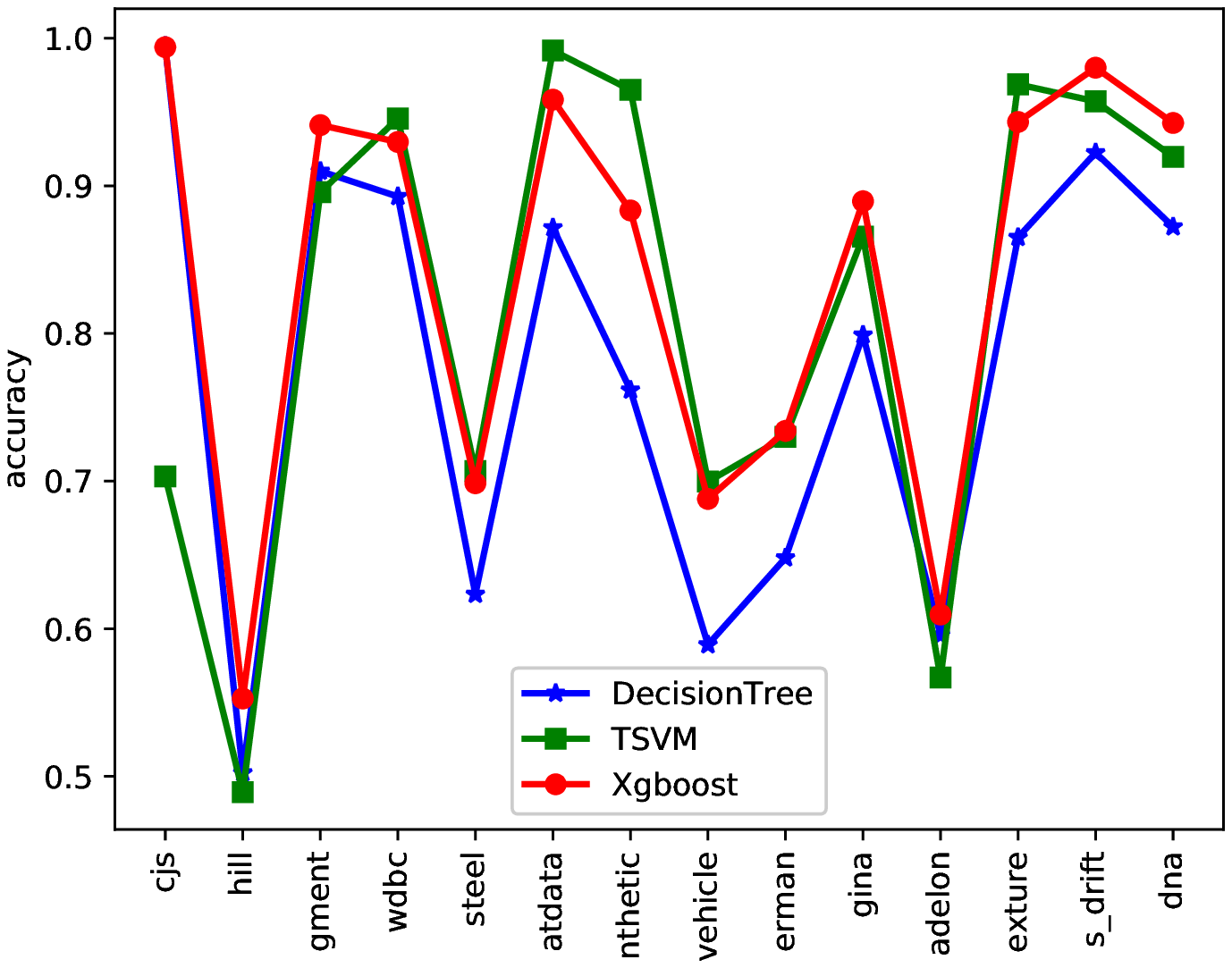}
\end{minipage}%
}%
\vspace{-0.2cm}
\centering
\caption{ (a) The correlation coefficient between two different classifiers; (b) The test accuracy of decision tree, TSVM and Xgboost.}
\label{correlation_accuracy}
\end{figure}

Fig. \ref{correlation_accuracy} (a) presents the diversity of two different classifiers. When the correlation coefficient $|\rho|$ is smaller, the diversity is larger. In \cite{li2007improve}, the authors proposed  Co-forest method, which uses some different decision tree as base learners, but we see that the correlation coefficient $\rho_{TT}$ between two different decision tree classifiers is higher than that between two heterogeneous classifiers. Especially, $\rho_{ST}$ and  $\rho_{SX}$ have much smaller value for \emph{analcat} data set and \emph{cjs} data set, which has been visualized in Fig. \ref{motivation}, it shows Xgboost and TSVM  have
complementary properties.

Fig. \ref{correlation_accuracy} (a) also shows $\rho_{SX}$ is bigger than $\rho_{ST}$, which means the diversity between TSVM and decision tree classifier is larger than that between TSVM and Xgboost. From Fig. \ref{correlation_accuracy} (b), however, it shows Xgboost always performs better than decision tree classifier. This phenomenon makes sense because Xgboost is an additive tree ensemble model. Compared with Xgboost, TSVM based on the large margin can obtain higher accuracy for some real data set, such as \emph{analcat} data. Thus, TSVM and Xgboost are selected as based learners  based on a tradeoff between diversity and accuracy.
\subsection{Initialization}\label{initia}
The new method, CTOW, comprises $K$ classifiers. Assume the $K$-th classifier is TSVM, and the others are Xgboost. In order to make CTOW  perform better for a real \emph{commercial} data, we choose several Xgboost rather than one Xgboost as base learner. In order to improve accurate and diverse classifiers further, each Xgboost is initiated from the different training sets bootstrapped from labeled set $\mathcal{L}$. For TSVM, the whole training set is used. Thus, we can train the $K$ classifiers simultaneously using the different data sets, which yields the prediction probability $P_k\in\mathcal{R}^{u\times C}$ of the unlabeled data, where $P_k=F_{\theta_k}(x_u)$, $k=1,\ldots,K$, and $\theta_k$ is the parameter for the $k$-th classifier.

\subsection{Optimally Combining Classifiers}\label{OWE}
Each classifier sends its prediction probability $P_k$ of the unlabeled data to the center, then the unlabeled data will be labeled by the optimal weight ensemble. Suppose $w_k$ is the weight of the $k$-th classifier. The probability of the pseudo label is denoted as
$
\hat{P}(\vw)=\sum_{k=1}^Kw_kP_k,
$
where $\vw$ is a vector with $\vw\triangleq[w_1,\ldots,w_K]$.

The optimal weight is obtained by solving the following optimization problem
\begin{align}
\label{opt1}\min_{\vw} & \frac{1}{u}\sum_{i=l+1}^{l+u}\sum_{j=1}^C-\hat{P}^{ij}(\vw)\log(\hat{P}^{ij}(\vw))+\mu\|\vw\|^2\\
\label{opt2}s.t.& \sum_{j=1}^Kw_j=1,w_j\geq 0,\\
\label{opt3} &w_K=h(\hat{\xi},\alpha),
\end{align}
where $\hat{P}^{ij}(\vw)$ means the prediction probability of the $i$-th unlabeled data from the $j$-th class. Recall that the $K$-th classifier is TSVM. We can give it a prior weight according to \eqref{opt3} with the following details. First, we introduce a margin density  metric for TSVM as follows \cite{sethi2017}:
\begin{align}
\label{hxi}\hat{\xi}=\frac{\#\{\xi_i>0,i=1,\ldots,l\}}{l}
\end{align}
where $\xi$ is the slack variable mentioned in \eqref{TSVM1}, the numerator counts the number of training samples that falls in the margin of the TSVM. The variable $\alpha$ is a threshold, and the goal of the function $h$ is to give a smaller weight to TSVM if $\hat{\xi}>\alpha$; otherwise it yields a bigger weight. In our experiment, the function $h$ is given as follows
\begin{align}
\label{h} h(\hat{\xi},\alpha) = \frac{1}{1+3\exp^{10(\hat{\xi}-\alpha)}},
\end{align}
where the threshold $\alpha$ will be specified shown in Section \ref{setup}.

The objective function in \eqref{opt1} contains the entropy of predicted probability and a regularization term $\|\vw\|^2$. We want to enforce  the ensemble classifier to provide low-entropy predictions on the unlabeled data. In addition, a regularization term is introduced to avoid overfitting to one classifier. When $\mu\rightarrow \infty$, the solution of the optimization problem \eqref{opt1}-\eqref{opt3} tends to give the same weight to all classifiers. The following example shows the relationship between \eqref{opt1}-\eqref{opt3} and the majority voting rule.

\textbf{Example 1.} Consider a binary classification problem. There are three classifiers to predict one instance with probability
$$
P_1=[1,0],P_2=[1,0],P_3=[0,1].
$$
Thus, we have
$$
\hat{P}(\vw)=[w_1+w_2,w_3]=[1-w_3,w_3].
$$
If we only minimize the entropy of $\hat{P}(\vw)$, then  there are infinite optimal solutions such as $\vw=[0,0,1]$ or $\vw=[a,1-a,0]$, where $a$ is an arbitrary constant. However, if we add the regularization term $\|\vw\|^2$, then a unique solution $\vw^{*}=[\frac{1}{2},\frac{1}{2},0]$ can be obtained with $\hat{P}(\vw^*)=[1,0]$, being equivalent to the result of applying majority voting rule.

Next, we show that the optimization problem \eqref{opt1}-\eqref{opt3} can be solved by projected gradient method. Specifically, by generating
the sequence $\{\vw^t\}$ via
\begin{align}
\label{pg} \vw^{t+1}=\Pi_X\left(\vw^t-\eta_t\nabla^t\right)
\end{align}
where the set $X$ is the linear constraint, i.e., $X=\{\vw:\sum_{j=1}^Kw_j=1,w_j\geq 0,w_K=h(\hat{\xi},\alpha)\}$, $\eta_t$ is the learning rate, $\nabla^t$ is the gradient of the objective function in \eqref{opt1} evaluated at $w^t$, and $\Pi_X(x)=\arg\min_{y\in X}\|x-y\|^2$ is Euclidean projection of $x$ onto $X$.  In this paper, we use an efficient algorithm proposed by  \cite{duchi2008efficient} to numerically perform the projection in \eqref{pg}.

\subsection{Diversity Augmentation}\label{DA}
In this subsection, we use the following steps to maintain diversity between base learners further.

Firstly, we utilize bootstrap sampling to select different subset $\mathcal{I}_k$ of the unlabeled data for the $k$-th classifier, where $|\mathcal{I}_k|=0.8u$.
Actually,  $\mathcal{I}_k$ can be also subsampled based on co-forest method proposed in \cite{li2007improve}, which can reduce the influence of misclassifying an unlabeled sample.

Secondly, inspired by the idea of the co-forest method, the probability $\bar{P}_k$ of the unlabeled data is obtained as follows
\begin{align}
\label{pseudo_label1} \bar{P}_k^{ij} =\sum_{s\neq k}^K w_s^{*} P_s^{ij},i\in \mathcal{I}_k, j\in \{1,\ldots,C\}
\end{align}
where $\vw^*$ is the optimal solution of problem \eqref{opt1}-\eqref{opt3}, $P_s^{ij}$ means the prediction probability of the $i$-th unlabeled data from the $j$-th class for the $s$-th classifier. If $F^*$ is the set containing all $K$ classifiers, the formulation in  \eqref{pseudo_label1} means all other component classifiers in $F^*$ without $F_{\theta_k}$ are used to determine
the most confidently unlabeled examples for the $k$-th classifier.
In order to filter out the unconfident pseudo labels, we select the unlabeled data $\bar{\mathcal{I}}_k$ from $\mathcal{I}_k$, when its maximum probability is bigger than a threshold $\beta$, i.e. $\bar{\mathcal{I}}_k=\{i:\max\limits_{j\in\{1,\ldots,C\}} \bar{P}_k^{ij}\geq \beta, i \in \mathcal{I}_k\}$.

Finally,  we get the reliable pseudo label for the $k$-th classifier as follows
\begin{align}
\label{pseudo_label2} \bar{y}_k^i = \arg\max_{j\in \{1,\ldots,C\}} \bar{P}_k^{ij}~ i \in \bar{\mathcal{I}}_k.
\end{align}
We then combine the pseudo labeled data $(x_i,\bar{y}_k^i)$ and labeled data $\mathcal{L}$ to train the $k$-th classifier again. The framework  of the whole training process is shown in Algorithm \ref{alg}. Based on the output of Algorithm \ref{alg}, we can predict the test data and calculate the test accuracy.
\begin{algorithm}[t]
\begin{algorithmic}[1]
\REQUIRE   Labeled set $\mathcal{L}$ and unlabeled set $\mathcal{U}$, $K$ classifiers, the threshold $\beta$ and $\alpha$, regularizer $\mu$, initializing weight $w^0$ and iterations $T$.
\FOR{$k=1:K$ (in parallel)}
\IF{$k\leq K-1$}
\STATE Learn the model function $F_{\theta_k}(\cdot)$ of Xgboost based on the $\mathcal{L}$.
\ELSE
\STATE Learn the model function $F_{\theta_k}(\cdot)$ of TSVM based on the $\mathcal{L}$ and $\mathcal{U}$.
\ENDIF
\ENDFOR
\REPEAT
\STATE Receive the probability $P_k=F_{\theta_k}(x_u)$ of unlabeled data, $k=1,\ldots,K.$
\FOR{$t=1:T$}
\STATE $w^{t+1}=\Pi_X\left(w^t-\eta_t\nabla^t\right)$
\ENDFOR
\STATE Achieve the optimal weight $w^{T+1}$ for problem \eqref{opt1}-\eqref{opt3}.
\FOR{$k=1:K-1$ (in parallel)}
\STATE Obtain pseudo label $ \bar{y}_k^i$ based on \eqref{pseudo_label1}-\eqref{pseudo_label2}, $i\in \bar{\mathcal{I}}_k$.
\STATE Augment the training sets $\bar{\mathcal{L}}_k=\mathcal{L}\cup \{x_i,\bar{y}_k^i\}_{i\in \bar{\mathcal{I}}_k}$
\STATE Learn the model function $F_{\theta_k}(\cdot)$ based on the $\bar{\mathcal{L}}_k$.
\ENDFOR
\UNTIL{none of the classifier changes}
\ENSURE Model parameter $\theta_k$ and optimal weight $w^{T+1}$.
\end{algorithmic}
\caption{Co-training with Optimal Weight}
\label{alg}
\end{algorithm}
\section{Experiments}\label{sec4}
For comparison, the performances of five state-of-the-art semi-supervised algorithms, i.e., Graph-SVM (GSVM, \cite{Belkin-Niyogi-Sindhwani06}), GMM \cite{zhu2005}, Ladder network \cite{rasmus2015}, Co-forest \cite{li2007improve}, TSVM \cite{bennett1999} are also evaluated.

 We conduct experiments based on 14 data sets from UCI machine learning repository. Additionally, we test our method on a real \textbf{commercial}
data set  which contains around 50 thousand samples with
62 features. Descriptions of the experimental data sets is shown in Table \ref{tab_data} (see Appendix).
\subsection{Implementation details}\label{setup}
For each data set, fivefold cross validation is employed for evaluation. For each fold, we split the training data in a stratified fashion to obtain a labeled data set $\mathcal{L}$ and an unlabeled set $\mathcal{U}$ for a given label rate $\gamma$. In our simulation, we set $\gamma=0.1$, which means that splitting the training set will produce a set with $10\%$ labeled examples and a set with $90\%$ unlabeled examples.

 The proposed method, CTOW, adaptively combines classification results of Xgboost and TSVM. In our simulation, we set $K=4$. Three Xgboost and one TSVM are used.  In addition, we set threshold $\beta=0.75$, iterations $T=10$ and penalty parameter $\mu=0.5$.

Firstly, let us show the relationship between margin density $\hat{\xi}$ (introduced in \eqref{hxi}) and the accuracy of both TSVM and Xgboost.
\begin{figure}[htbp]
\centering
\subfigure[]{
\hspace{-0.7cm}
\begin{minipage}[t]{0.5\linewidth}
\centering
\includegraphics[height=2in]{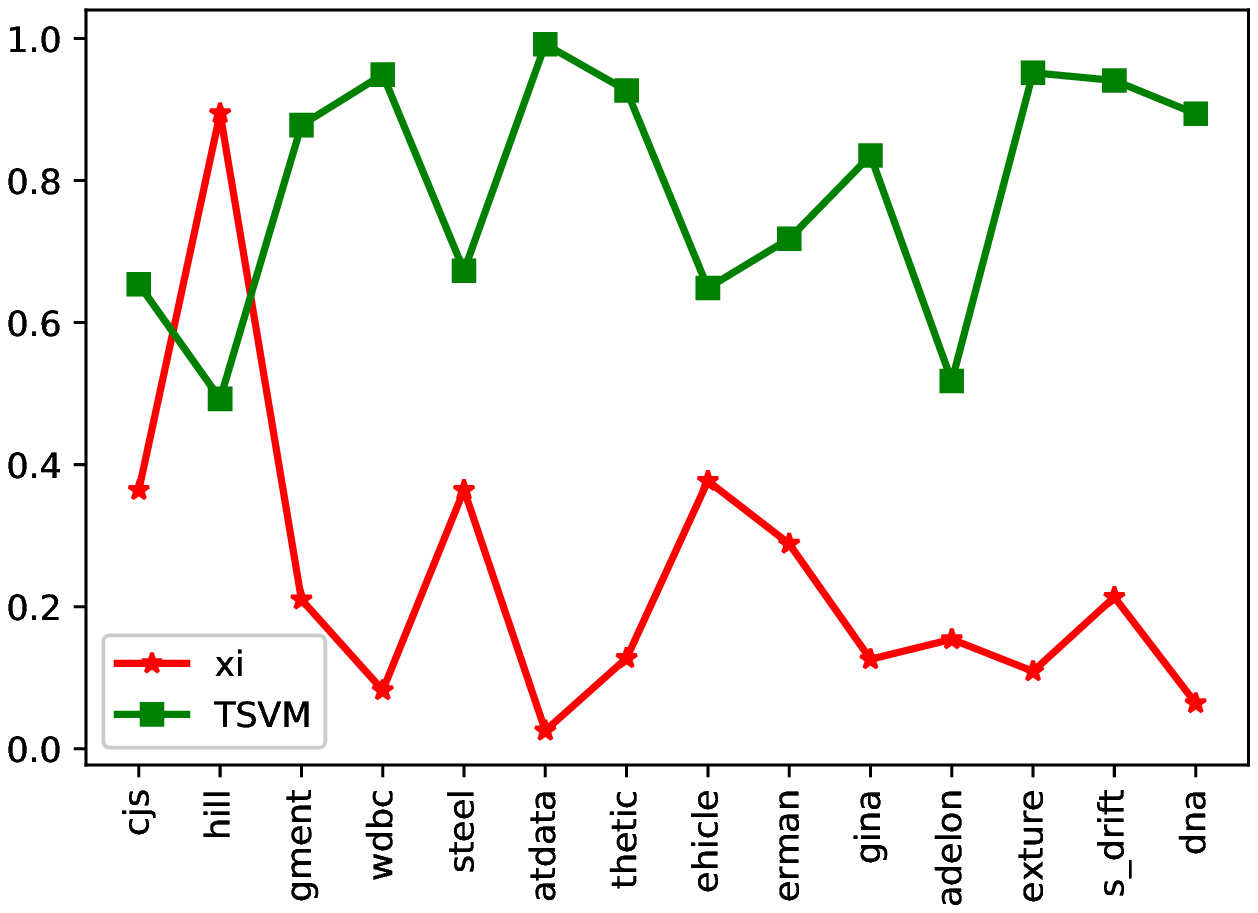}
\end{minipage}%
}%
\subfigure[]{
\hspace{-0.12cm}
\begin{minipage}[t]{0.5\linewidth}
\centering
\includegraphics[height=2in]{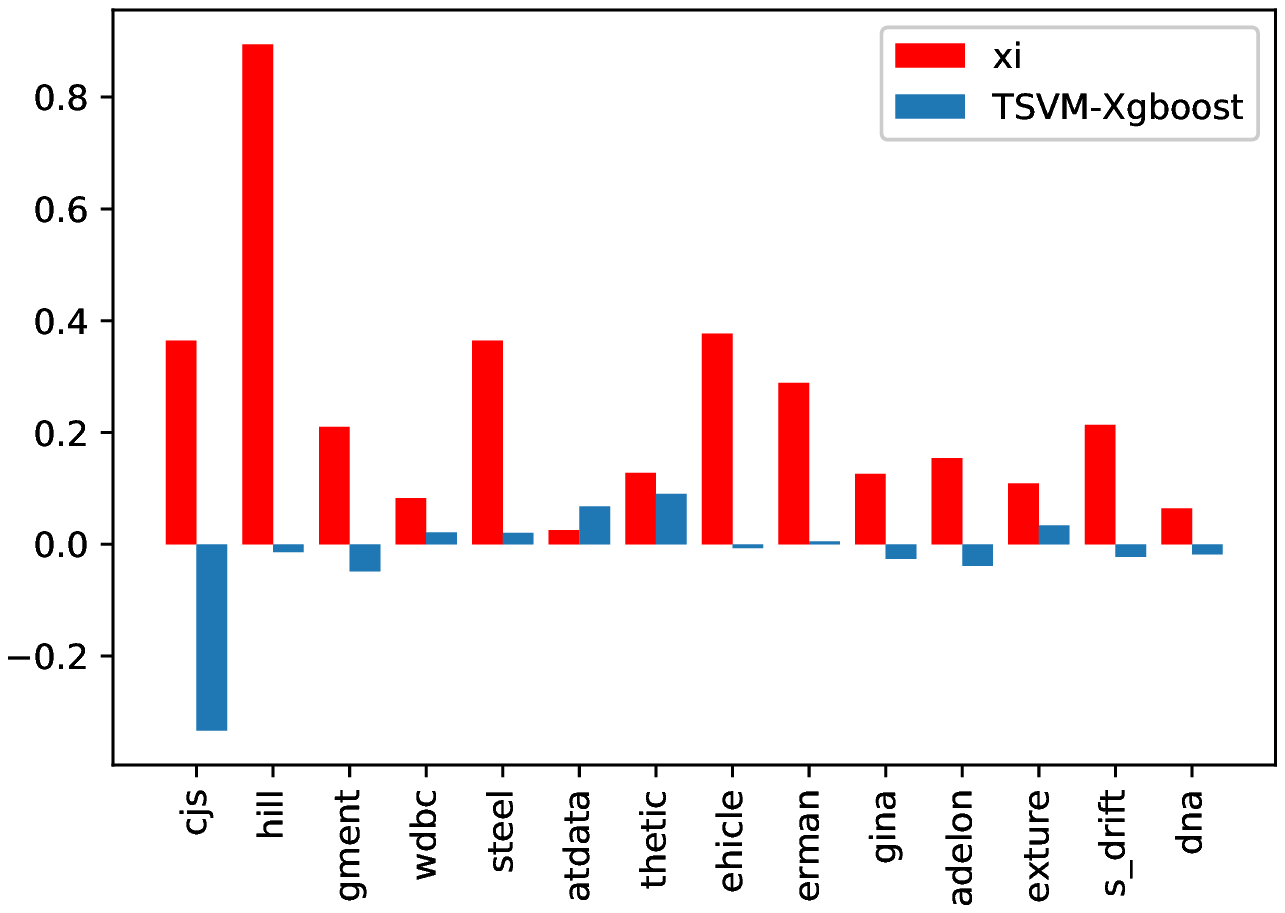}
\end{minipage}%
}%
\vspace{-0.2cm}
\centering
\caption{(a): The relationship between the accuracy of TSVM and $\hat{\xi}$; (b) The accuracy of TSVM minus that of Xgboost, and compare it with $\hat{\xi}$.}
\label{TSVM_xi}
\end{figure}
Fig. \ref{TSVM_xi} (a) shows that when $\hat{\xi}$ is smaller, the accuracy of TSVM tends to be higher. The bar with negative value in Fig. \ref{TSVM_xi} (b) means that Xgboost performs better than TSVM, which happens when $\hat{\xi}$ becomes larger (see \emph{cjs} data).  Then, $\hat{\xi}$ can be used as the prior information for providing the weight of TSVM. Specifically, if $\hat{\xi}$ is bigger than a threshold $\alpha$, we give a smaller weight to TSVM, or a larger weight otherwise. According to the above discussion and observation from Fig. \ref{TSVM_xi},   we set $\alpha=0.2$ in our Algorithm \ref{alg}, and use the function $h$ denoted in \eqref{opt3} to calculate the weight $w_k$ of TSVM.
Based on the above setup, we run Algorithm \ref{alg} to show the results in the next subsection.
\subsection{Performance}\label{sub_result}
\begin{figure}[htbp]
\centering
\subfigure[commercial]{
\hspace{-0.6cm}
\begin{minipage}[t]{0.5\linewidth}
\centering
\includegraphics[height=2.1in]{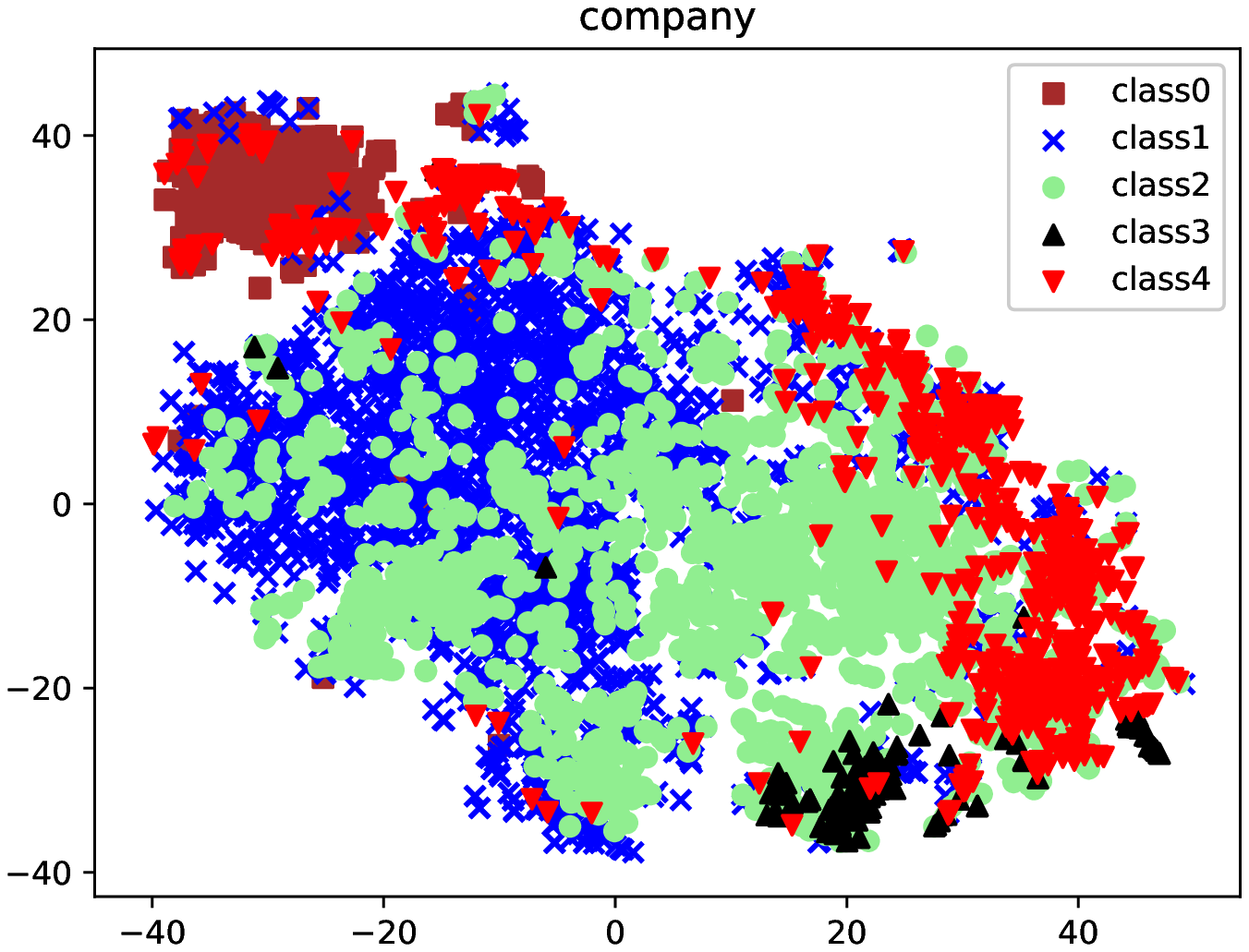}
\end{minipage}%
}%
\subfigure[synthetic]{
\hspace{-0.6cm}
\begin{minipage}[t]{0.5\linewidth}
\centering
\includegraphics[height=2.1in]{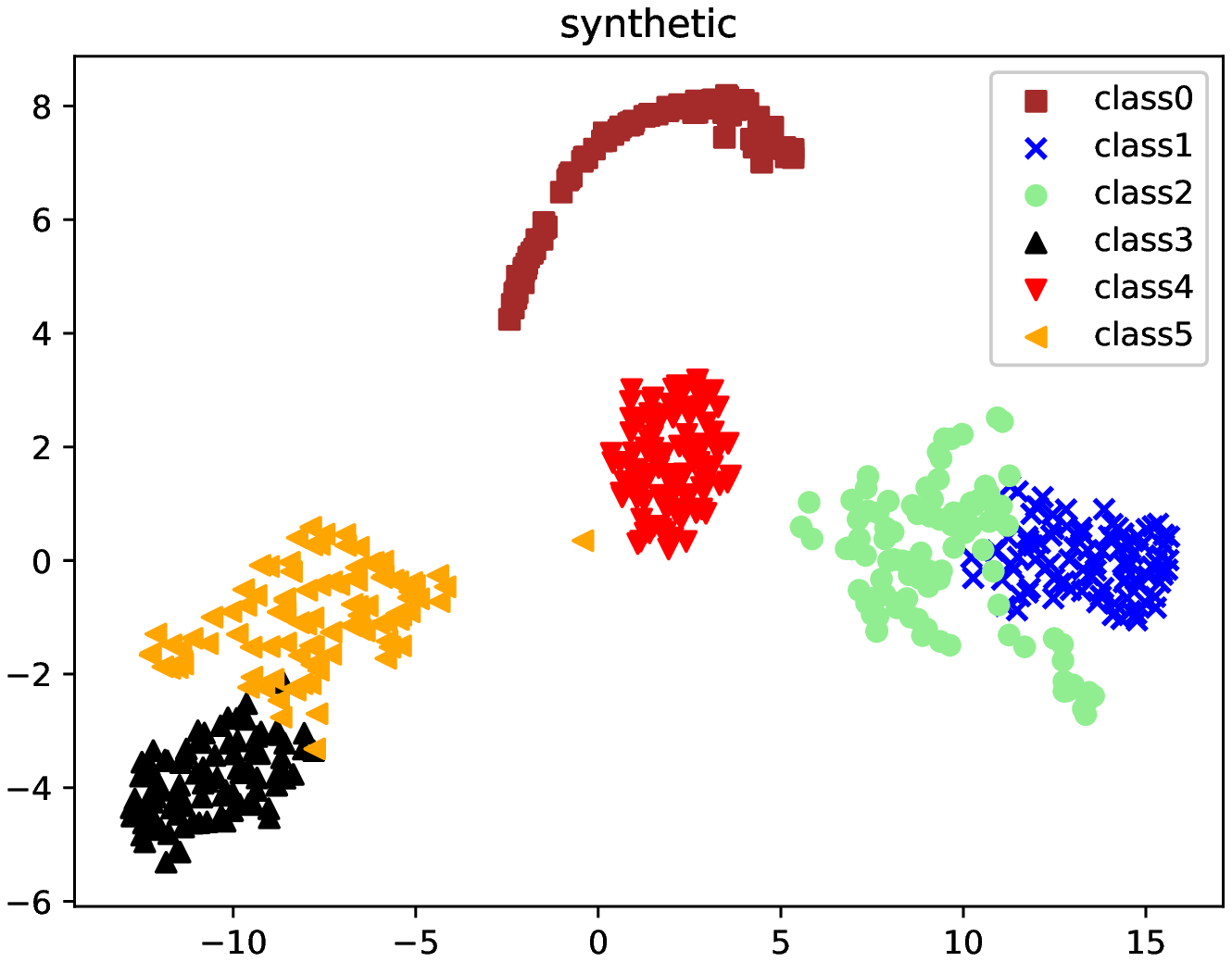}
\end{minipage}%
}%
\centering
\vspace{-0.2cm}
\caption{Visualization of the real data.}
\label{Vis2}
\end{figure}

Firstly, the visualization of some data sets is presented in Fig. \ref{motivation} and Fig. \ref{Vis2}, which helps us to analyze the suitable application for these methods.  Secondly, we compare the proposed method with some other semi-supervised learning methods, which shows that our method yields the best performance in many real data sets.
 Table \ref{test_acc} shows the test accuracy of the different algorithms. From the second column, we see that GMM fails to work well in various data sets. The reason is that GMM needs to assume the data comes from Gaussian mixture distribution, which is a strong assumption for real data sets. Comparing Graph-SVM with TSVM, Table \ref{test_acc} presents that these two methods have similar performance, because both of them are based on SVM.
  Nevertheless, Graph-SVM needs to calculate the similarity matrix, then it takes too long time for computation (see Fig \ref{top-ave1}(a)). Thus, we do not report the result of GMM and Graph-SVM if the data size is large.

In Fig. \ref{Vis2}, \emph{synthetic} data enjoys large margin so that TSVM performs better than Co-forest (see
 Table \ref{test_acc}). Conversely,  the classes of both \emph{commercial} and \emph{cjs} data sets are overlapped and the data distribution is irregular, then co-forest based tree model can obtain better performance than TSVM. Ladder network can get the best performance for \emph{hill} and \emph{texture} data sets, because these original features have homogeneous attributes and they cannot represent the label very well. But the neural network can train a better feature space in this case. Our method utilizes the advantage of Xgboost and TSVM.  Table \ref{test_acc} presents that the proposed method can achieve the best performance for many real data sets, especially for large and high dimension data, such as \emph{gas-grift} and \emph{commercial} data set (see Table \ref{tab_data}).
\begin{table}[t]
\centering
\caption{Test accuracy of the compared algorithms}
\label{test_acc}
\setlength{\tabcolsep}{2pt}
\begin{tabular}{ccccccc}
\hline
Data& GMM& GSVM&Ladder&Co-forest&TSVM&CTOW\\
\hline
cjs& 0.293& 0.640 & 0.740&\textbf{0.989}&0.654&0.987\\
hill&0.488 & 0.490 & \textbf{0.530} &0.492&0.493&0.499\\
segment&0.694 & 0.889 & 0.898&0.907&0.878&\textbf{0.925}\\
wdbc&0.643 &0.940&0.932&0.905&0.949&\textbf{0.954}\\
steel& 0.466& 0.627&0.652&0.620&\textbf{0.673}&0.649\\
analcat&0.206& 0.975&	0.982&	0.876&	0.992&	\textbf{0.993}\\
synthetic& 0.292&	0.908&	0.810&	0.745&	\textbf{0.927}&	0.920\\
vehicle&\textbf{0.657}&0.596&0.635&0.631&0.649&0.625\\
german&0.614&0.619&0.679&0.712&\textbf{0.718}&0.716\\
gina&*&*&0.807&0.814&0.835&\textbf{0.857}\\
madelon&*&*&0.536&0.538&0.518&\textbf{0.543}\\
texture&*&*&\textbf{0.973}&0.877&0.952&0.953\\
gas-grift&*&*&0.945&0.927&0.941&\textbf{0.965}\\
dna&*&*&0.885&0.890&0.894&\textbf{0.911}\\
\textbf{commercial}&*&*&0.832&0.816&0.861&\textbf{0.901}\\
\hline
\end{tabular}
\end{table}
\section{Conclusions and Future Works}\label{sec5}
In this paper, we propose a new method, CTOW, for semi-supervised deep
learning,  which applies the optimal ensemble of two heterogeneous classifiers, namely  Xgboost and  TSVM.  The unlabeled
data is exploited by considering model initialization, solving optimal weight ensemble problem, diversity augmentation, simultaneously. Experiments on various real data sets  demonstrate that our method is superior to state-of-the-art methods.

From the simulations, we find Xgboost and TSVM have  complementary properties and larger diversity. The reason that leading to this phenomenon should be studied in the future.
\section*{Broader Impact}
Although SSL is an old topic, there has been renewed interest in SSL which is reflected in both academic and industrial research. Many SSL methods are proposed based on certain assumptions on the labeled data and unlabeled data, which plays an important role in semi-supervised learning.  However, it remains an open question on how to make the right assumptions on a real data.

Our research finds Xgboost and TSVM have  complementary properties and larger diversity. Then,
we proposed a new SSL method called CTOW by appling the optimal ensemble of two heterogeneous classifiers Xgboost and  TSVM. Thus, the proposed CTOW enjoys the advantage of both Xgboost and TSVM such that it weakly depends on the distribution of the training data.

Our research could be also used to provide explanations for Xgboost and TSVM in their applications as well as reducing the cost of labeling. The proposed method may fail when label data is very few such that both Xgboost and TSVM performs worst.

\bibliographystyle{unsrt}
\bibliography{wang_bibtex}
\newpage
\section{Appendix}
\subsection{Data sets}
Detailed information of the experimental data sets is shown in Table \ref{tab_data}.
\begin{table}[htbp]
\centering
\caption{Experimental data sets}
\label{tab_data}
\begin{tabular}{cccc}
\hline
Data set& instances& feature&classes\\
\hline
cjs& 2796 & 10&6\\
hill & 1212 & 101&2\\
segment & 2310 & 20&7\\
wdbc &569&31&2\\
steel&1941&27&7\\
analcat&841&71&4\\
synthetic& 600&62&7\\
vehicle&846&19&4\\
german&1000&24&2\\
gina&3468&971&2\\
madelon&2600&500&2\\
texture&5500&41&11\\
gas-grift&13910&129&6\\
dna&3186&181&3\\
\textbf{commercial}&50000&62&5\\
\hline
\end{tabular}
\end{table}
\subsection{Visualization}
\begin{figure}[htbp]
\centering
\subfigure[madelon]{
\hspace{-0.15cm}
\begin{minipage}[t]{0.5\linewidth}
\centering
\includegraphics[height=2in]{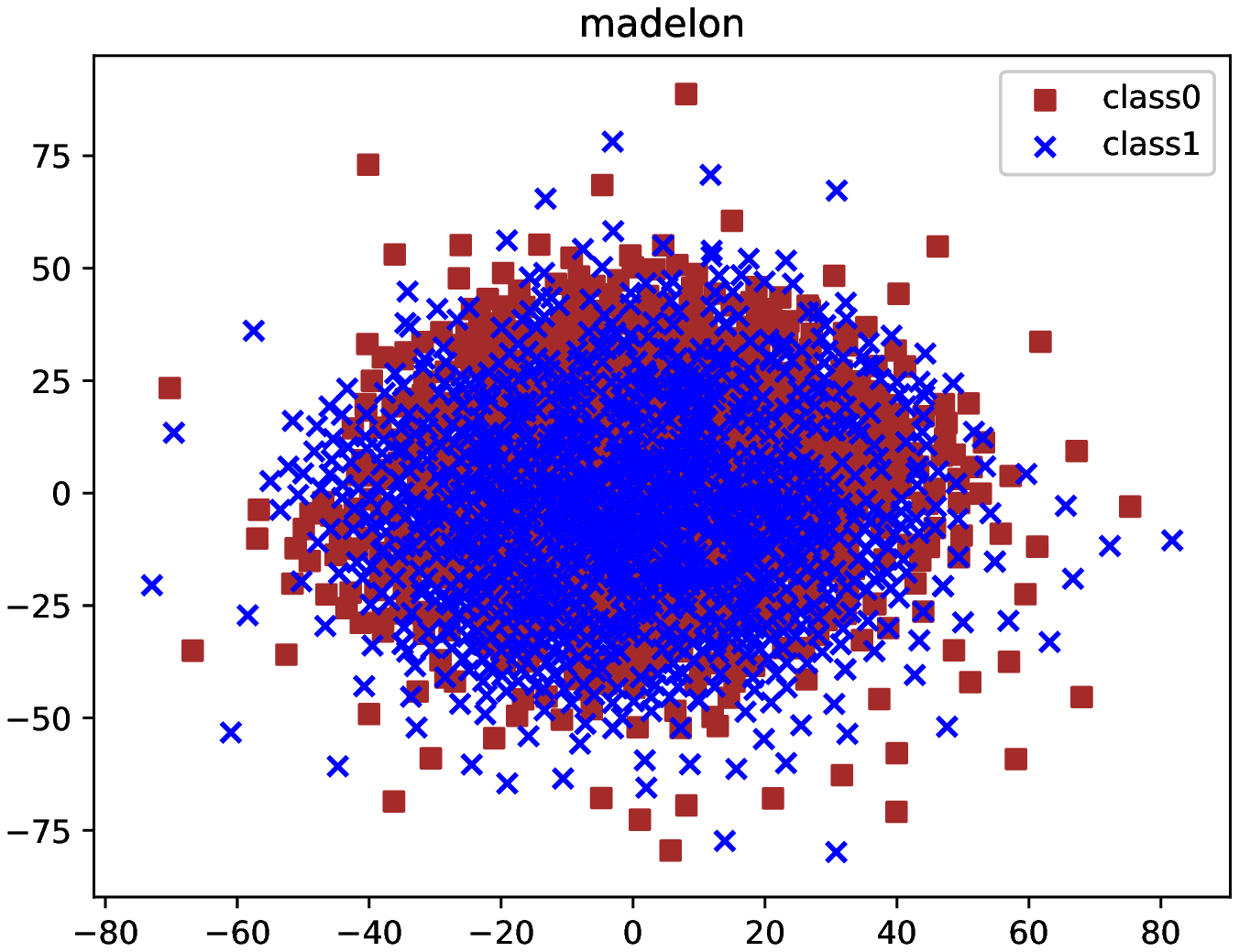}
\end{minipage}%
}%
\subfigure[texture]{
\hspace{-0.15cm}
\begin{minipage}[t]{0.5\linewidth}
\centering
\includegraphics[height=2in]{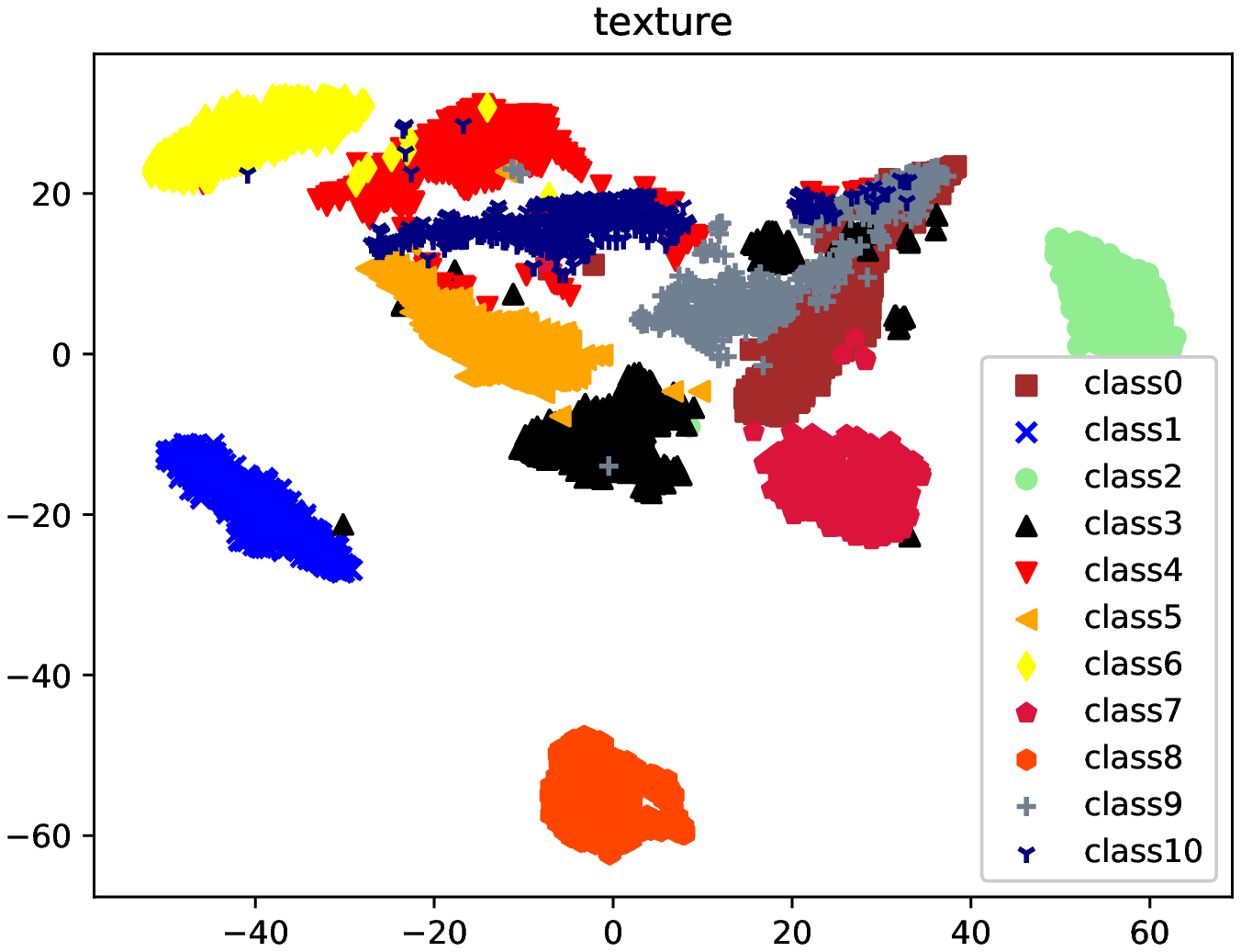}
\end{minipage}%
}%

\subfigure[segment]{
\hspace{-0.15cm}
\begin{minipage}[t]{0.5\linewidth}
\centering
\includegraphics[height=2in]{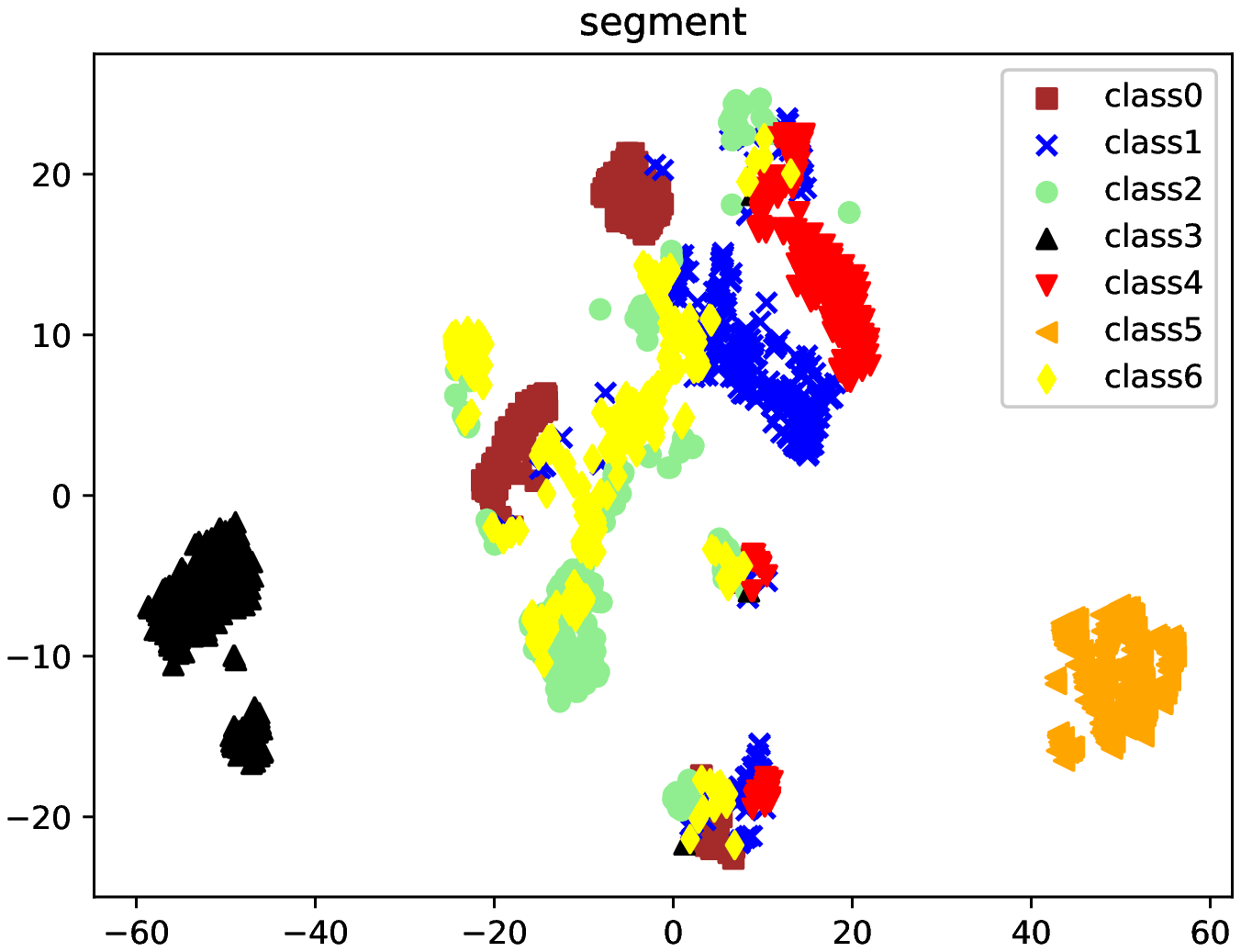}
\end{minipage}%
}%
\subfigure[wdbc]{
\hspace{-0.15cm}
\begin{minipage}[t]{0.5\linewidth}
\centering
\includegraphics[height=2in]{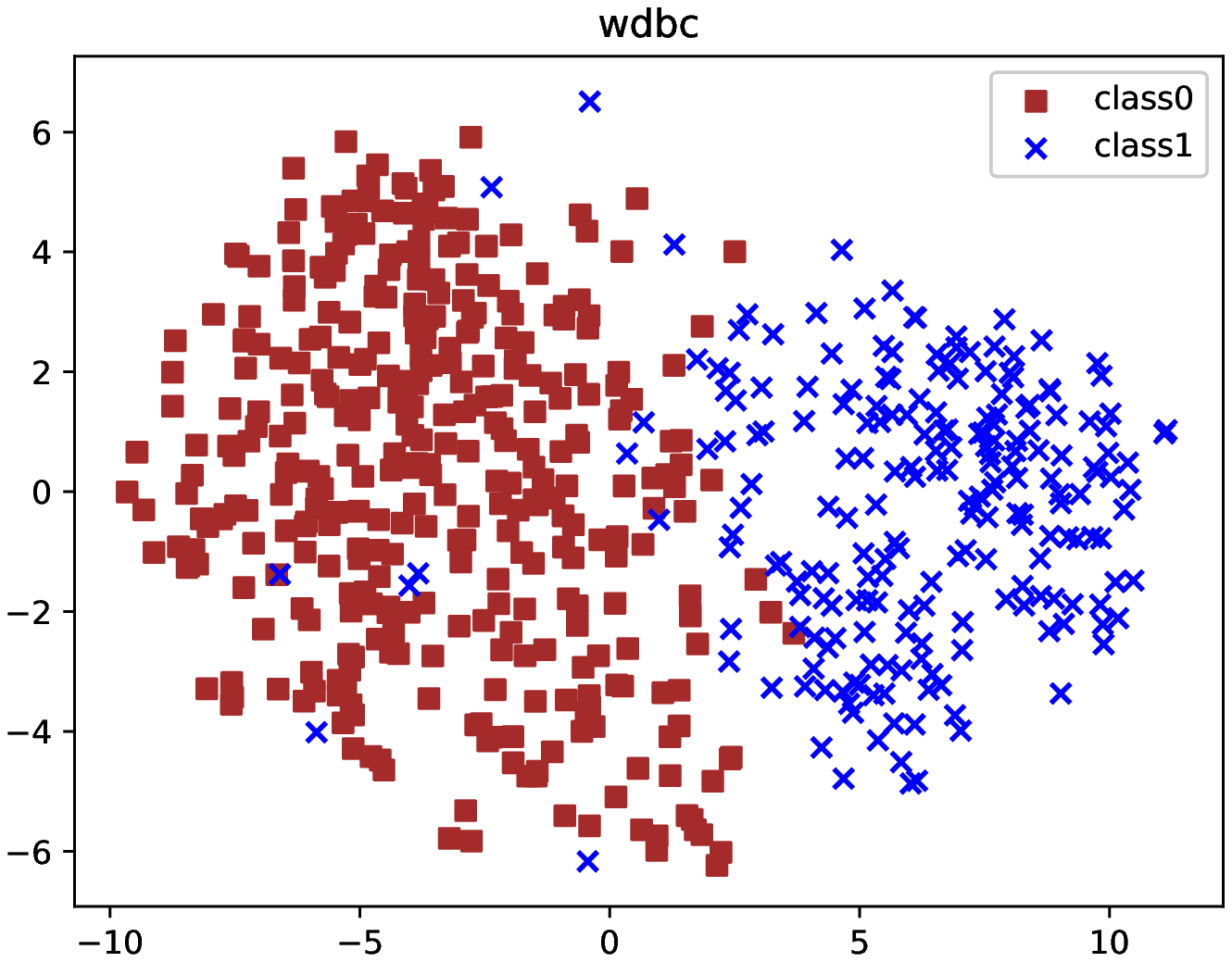}
\end{minipage}%
}%
\centering
\vspace{-0.2cm}
\caption{Visualization of the real data.}
\label{Vis3}
\end{figure}

The visualization of some other real data sets are presented in Fig. \ref{Vis3}, which helps us to select the suitable application for these methods.
\subsection{Performance Analysis}
\begin{figure}[t]
\centering
\subfigure[]{
\hspace{-0.25cm}
\begin{minipage}[t]{0.5\linewidth}
\centering
\includegraphics[height=1.8in]{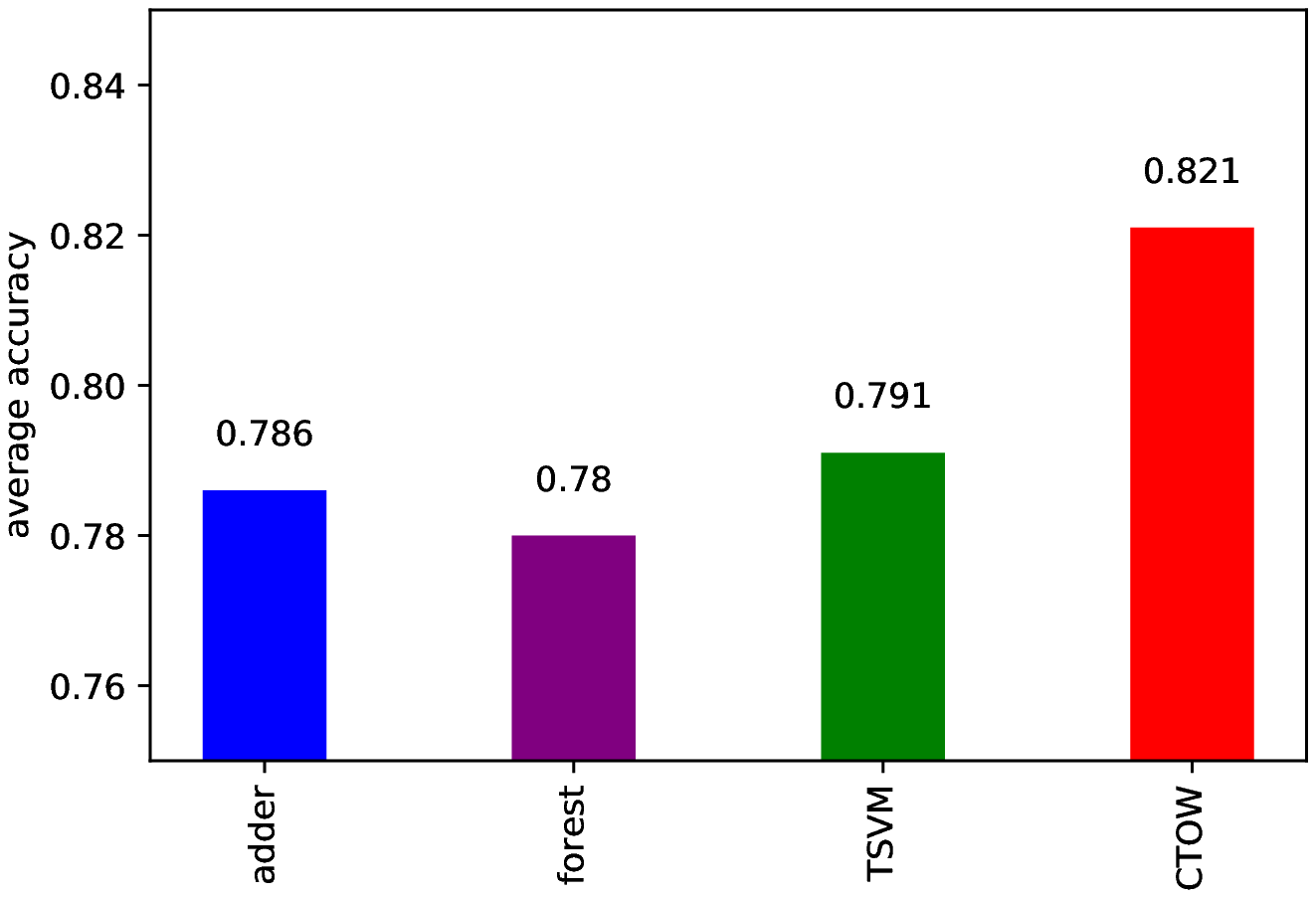}
\end{minipage}%
}%
\subfigure[]{
\hspace{-0.9cm}
\begin{minipage}[t]{0.5\linewidth}
\centering
\includegraphics[height=1.8in]{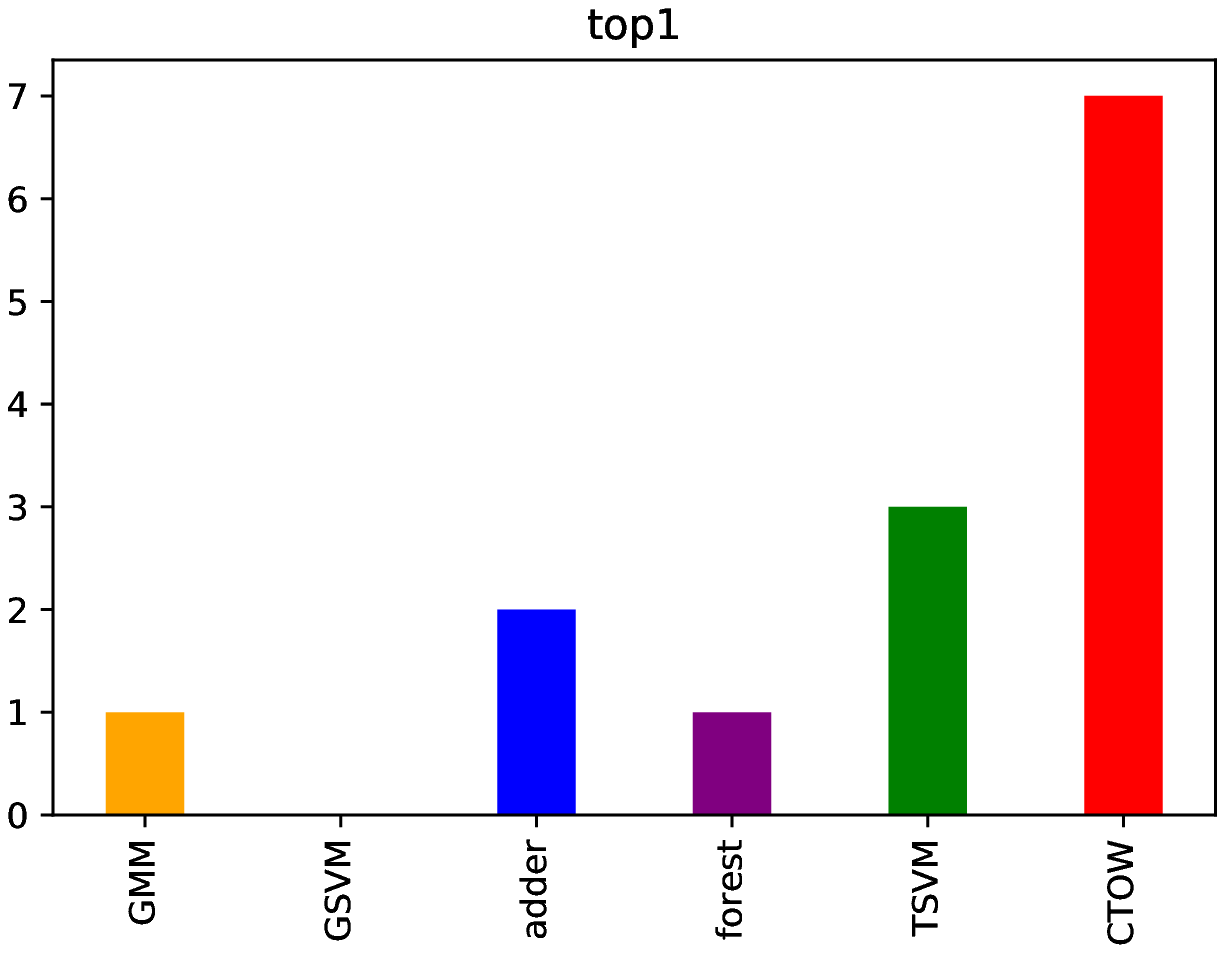}
\end{minipage}%
}%
\centering
\vspace{-0.2cm}
\caption{(a): The average accuracy of different algorithms with 14 data sets; (b) Count the number of times that different methods reach the highest accuracy.}
\label{top-ave}
\end{figure}

Fig. \ref{top-ave} (a) presents the average accuracy of different algorithms with 14 data sets, the proposed method, CTOW, improves at least $\%3$ accuracy comparing with the other methods. It is well known that one algorithm cannot always beat the other methods, but we can count the number of times that these algorithms achieves the highest test accuracies in Fig. \ref{top-ave} (b). It shows that our proposed method can achieve the best performance for half of these real data sets, which contains large margin and irregular data.

\begin{figure}[t]
\centering
\subfigure[]{
\begin{minipage}[t]{0.5\linewidth}
\centering
\includegraphics[height=2in]{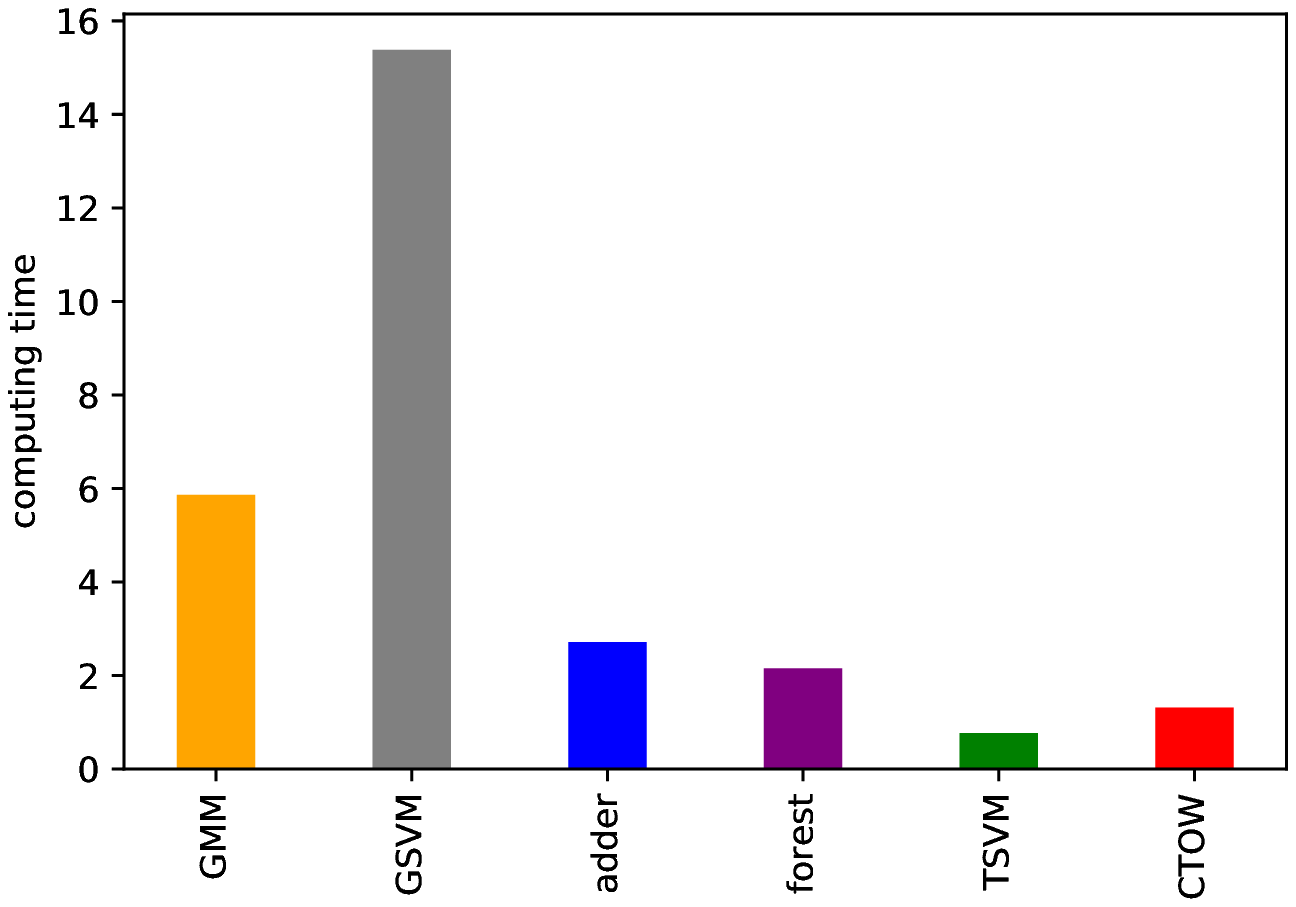}
\end{minipage}%
}%
\subfigure[]{
\hspace{-0.25cm}
\begin{minipage}[t]{0.5\linewidth}
\centering
\includegraphics[height=2in]{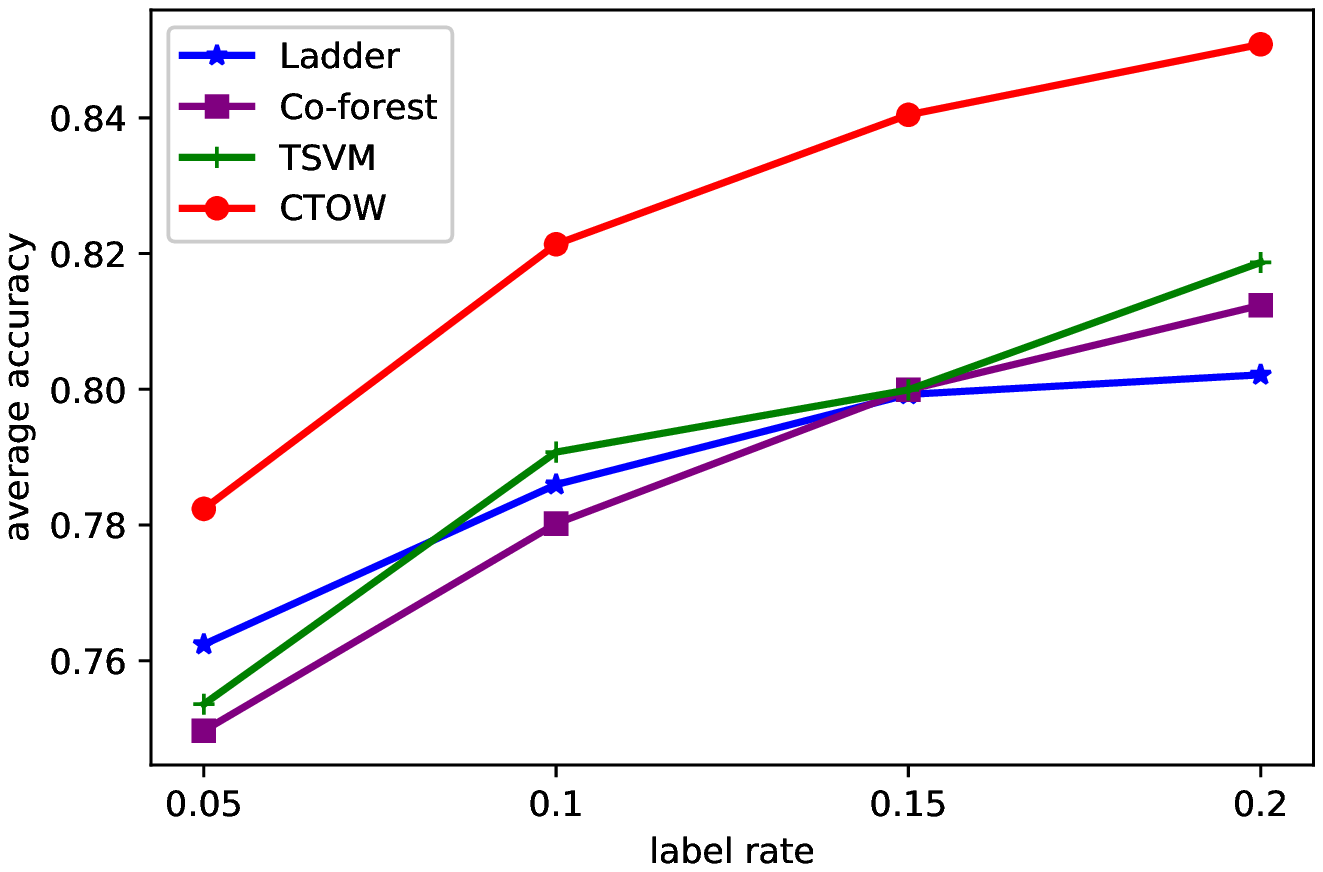}
\end{minipage}%
}%
\centering
\vspace{-0.2cm}
\caption{(a): The square root of  computing time of different methods for cjs data; (b) The average accuracy of different algorithms with the label rate $\gamma$.}
\label{top-ave1}
\end{figure}

 Fig. \ref{top-ave1} (b) presents the average accuracy of different algorithms with 14 data sets with different
 label rate $\gamma$, the proposed method, CTOW, improves at least $3\%$ accuracy comparing with the other methods
 for label rate $\gamma=0.1$. With the increase of the label rate $\gamma$, our method performs better than Co-forest, Ladder network and TSVM.
\subsection{Ablation Study}
Since our method combines several semi-supervised learning methods, here, we show an
ablation study and discuss the effect of removing some components in order to provide additional insight about the proposed method.
Specifically, we measure the performance of CTOW without considering prior information of margin density in \eqref{opt3}, which is denoted as CTOW-NP. Removing the classifier, TSVM, and only use co-training with Xgboost, which is called CTOW-NT.

Table \ref{Ablation} summarizes our ablation results. It shows that only using Xgboost or TSVM degrades the classifier's performance. Meanwhile, correct prior information of the margin density can help us obtain improved performance.
\begin{table}[t]
\centering
\caption{Ablation study}
\label{Ablation}
\setlength{\tabcolsep}{3pt}
\begin{tabular}{cccc}
\hline
Data&CTOW-NP&CTOW-NT&CTOW\\
\hline
cjs&0.987&0.989&0.987\\
hill &	0.499&	0.502&	0.499\\
segment&	0.923&	0.922&	\textbf{0.925}\\
wdbc &	0.931&	0.919&	\textbf{0.954}\\
steel&	0.647&	0.646&	\textbf{0.649}\\
analcat&	0.961&	0.924&	\textbf{0.993}\\
synthetic&	0.898&	0.815&	\textbf{0.920}\\
vehicle&	0.653&	0.657&	0.625\\
german&	0.715&	0.709&	\textbf{0.716}\\
gina&	0.858&	0.864&	0.857\\
madelon &	0.562&	0.556&	0.543\\
texture&	0.931&	0.915&	\textbf{0.953}\\
gas-grift&	0.962&	0.964&	\textbf{0.965}\\
dna&	0.911&	0.912&	0.911\\
\hline
Average&	0.817&	0.807&	0.821\\
\hline
\end{tabular}
\end{table}
\end{document}